\documentclass[sigconf]{acmart}

\setcopyright{cc}
\setcctype{by}
\copyrightyear{2025}

\acmConference[MM '25]{Proceedings of the 33rd ACM International Conference on Multimedia}{October 27--31, 2025}{Dublin, Ireland}
\acmBooktitle{Proceedings of the 33rd ACM International Conference on Multimedia (MM '25), October 27--31, 2025, Dublin, Ireland}
\acmDOI{10.1145/3746027.3754749}
\acmISBN{979-8-4007-2035-2/2025/10}

\AtBeginDocument{%
  }
\usepackage{colortbl}
\usepackage{multirow}

\usepackage{amsmath}
\usepackage{algorithm}
\usepackage{algorithmic}
\usepackage{bm}
\usepackage{makecell}
\usepackage{tikz}
\usetikzlibrary{positioning, arrows.meta}
\begin{document}

\acmSubmissionID{432}




\title{EgoPrompt: Prompt Learning for Egocentric Action Recognition}


\author{Huaihai Lyu}
\affiliation{%
  \department{MAIS, Institute of Automation,}
  \institution{Chinese Academy of Sciences}
  \city{Beijing}
  \country{China}}
\email{lvhuaihai2023@ia.ac.cn}

\author{Chaofan Chen}
\affiliation{%
  \department{MAIS, Institute of Automation,}
  \institution{Chinese Academy of Sciences}
  \city{Beijing}
  \country{China}}
\email{chencfbupt@gmail.com}
\authornote{Corresponding author.}

\author{Yuheng Ji}
\affiliation{%
  \department{MAIS, Institute of Automation,}
  \institution{Chinese Academy of Sciences}
  \city{Beijing}
  \country{China}}
\email{jiyuheng2023@ia.ac.cn}

\author{Changsheng Xu}
\affiliation{%
  \department{MAIS, Institute of Automation,}
  \institution{Chinese Academy of Sciences and Peng Cheng Laboratory}
  \city{Beijing}
  \country{China}}
\email{csxu@nlpr.ia.ac.cn}
\renewcommand{\shortauthors}{Huaihai Lyu, Chaofan Chen, Yuheng Ji, \& Changsheng Xu}






\begin{abstract}
Driven by the increasing demand for applications in augmented and virtual reality, egocentric action recognition has emerged as a prominent research area. 
It is typically divided into two subtasks: recognizing the performed behavior (i.e., \textit{verb component}) and identifying the objects being acted upon (i.e., \textit{noun component}) from the first-person perspective. 
However, most existing approaches treat these two components as independent classification tasks, focusing on extracting component-specific knowledge while overlooking their inherent semantic and contextual relationships, leading to fragmented representations and sub-optimal generalization capability.
To address these challenges, we propose a prompt learning-based framework, \textit{\textbf{EgoPrompt}}, to conduct the egocentric action recognition task.
Building on the existing prompting strategy to capture the component-specific knowledge, we construct a \textit{\textbf{Unified Prompt Pool}} space to establish interaction between the two types of component representations.
Specifically, the component representations (from verbs and nouns) are first decomposed into fine-grained patterns with the prompt pair form.
Then, these pattern-level representations are fused through an attention-based mechanism to facilitate cross-component interaction.
To ensure the prompt pool is informative, we further introduce a novel training objective, \textit{\textbf{Diverse Pool Criteria}}.
This objective realizes our goals from two perspectives: \textit{Prompt Selection Frequency Regularization} and \textit{Prompt Knowledge Orthogonalization}.
Extensive experiments are conducted on the Ego4D, EPIC-Kitchens, and EGTEA datasets. 
The results consistently show that EgoPrompt achieves state-of-the-art performance across within-dataset, cross-dataset, and base-to-novel generalization benchmarks.
\end{abstract}
\maketitle

\begin{figure}
    \centering
    \includegraphics[width=1.0\columnwidth]{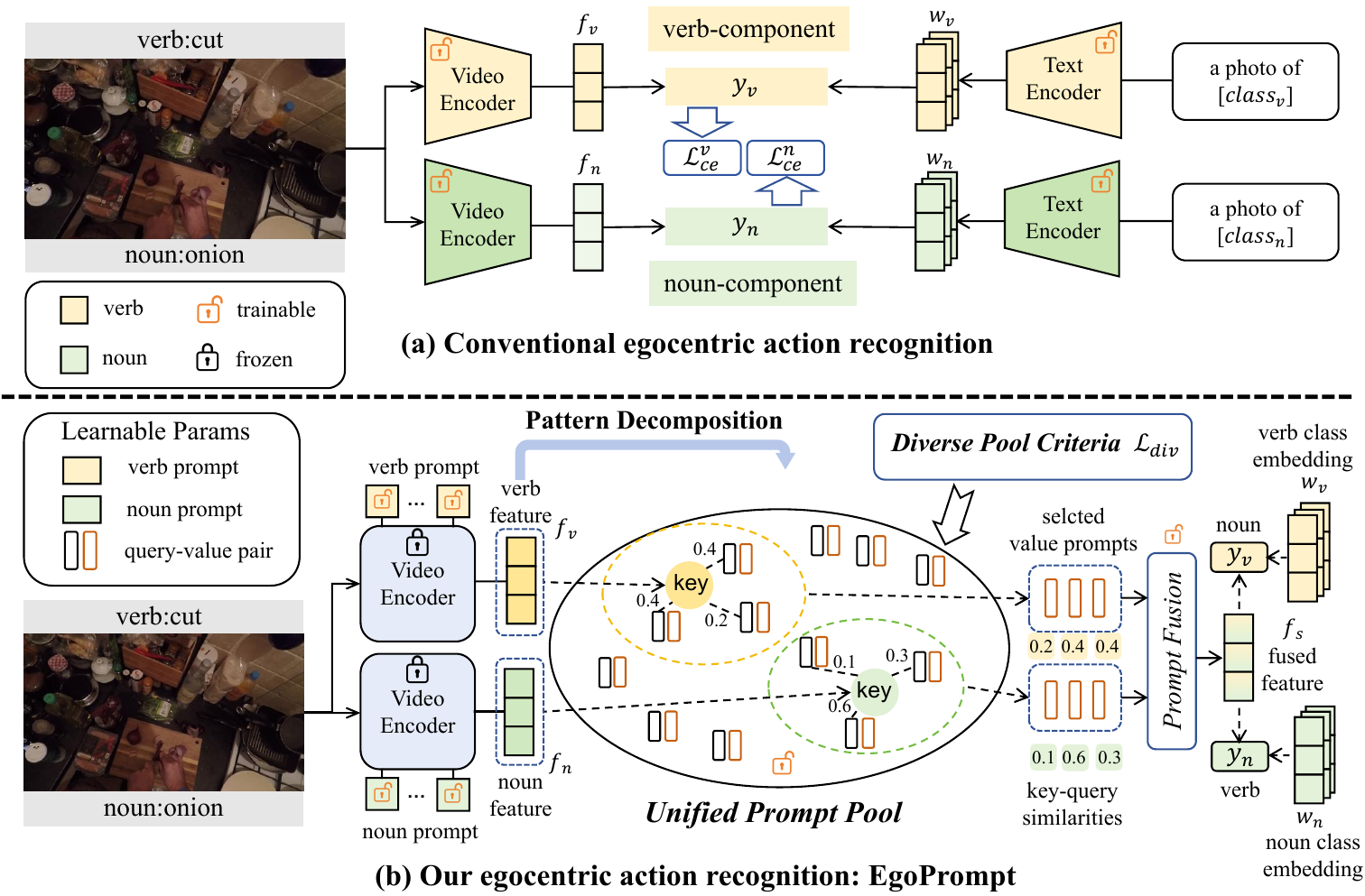}
    \caption{Comparison with existing framework.
    (a) Conventional methods use the component label
    of verb/noun to fine-tune its corresponding encoder independently. 
    (b) Our proposed EgoPrompt constructs a \textit{Unified Prompt Pool} with \textit{Diverse Pool Criteria} constraint, which decomposes component-specific representation into implicit prompt pair patterns and achieves better knowledge interaction with an attention fusion mechanism. 
    }
    \label{intro}
   \vspace{-1.1em}
\end{figure}
\section{Introduction}\label{sec:intro}
Recent advancements in augmented and virtual reality (AR/VR) technologies~\cite{llava-interleave, llava-ov} have demonstrated the potential to transform the way humans interact with the digital world. 
A fundamental requirement for such AR/VR systems is the ability to recognize user behaviors (\textit{verb component}) and the objects they interact with (\textit{noun component}) from egocentric videos.  
This demand has spurred a growing interest in egocentric action recognition (EAR)~\cite{egodistill,shiota2024egocentric,helping_hands}, which focuses on understanding first-person visual data to enable applications such as assistive systems and wearable devices.

Although existing EAR methods~\cite{AoP,xu2023pov} have made significant progress, their generalization performance remains insufficient when adapting to real-world datasets with distribution shifts.
Such distribution shifts arise from variations in the environment, recording conditions, and user behaviors, posing significant challenges for robust egocentric video understanding~\cite{lavila, egtea, ego4d, epic-kitchen, huang2024egoexolearn}. 
To address this issue, researchers have introduced some parameter-efficient strategies~\cite{chen2025pseudo}, such as prompt learning~\cite{VPT, tcp, khattak2023maple} and adapter-based techniques~\cite{doprompt,clipadapter}, to effectively adapt to the EAR tasks. 
Unlike third-person perspectives that require a comprehensive understanding of the entire environmental context, the semantic content of egocentric video is primarily reflected in the Human Object Interaction (HOI) region~\cite{AoP}.
However, the visual field in egocentric videos is often cluttered with \textit{irrelevant objects and background noise} in the HOI region, which significantly complicates the action recognition task. 
As shown in Fig.~\ref{intro} (a), most existing methods treat verbs and nouns as separate entities to capture component-specific knowledge.
These methods neglect the strong relationship between behaviors and the objects being acted upon in EAR.
For example, the interactive object ``carrot'' could constrain the semantic space of verbs (\emph{e.g.,} ``slicing'' and ``cutting'') depending on the context, and the verb “cutting” has specified the object of interaction with attributions (\emph{e.g.,} ``cuttable'' and ``solid'').
This interdependence highlights that they are not independent components but instead mutually influence one another, which collectively reflects the egocentric semantic content and helps in identifying HOI information from the noisy environment.

Based on the above analysis, we propose a novel prompt learning-based framework, \textit{\textbf{EgoPrompt}}, to explore the nature of semantic interplay in EAR task.
Built upon the component-specific feature extraction, we further construct a \textbf{\textit{Unified Prompt Pool}} space that encodes fine-grained implicit patterns through multiple query-value prompt pairs. 
Specifically, as illustrated in Fig~\ref{intro} (b), the component-specific representations serve as keys that match with queries in the prompt pool. 
Matched queries retrieve corresponding value prompts, representing the decomposed latent patterns from the input representation.
Finally, the selected value prompts are integrated into a fused representation that facilitates cross-component semantic interaction. 
By enabling the verb and noun features to interact with pattern prompt pairs through attention-based fusion, the egocentric model effectively captures their contextual semantic interplay. 
This interaction facilitates the integration of complementary information from both components, leading to a more context-aware understanding of HOI semantics.
To ensure the Unified Prompt Pool captures the informative pattern knowledge, we introduce a novel training objective, {\textit{\textbf{Diverse Pool Criteria}}, to consider the following two factors:
(1) \textit{Prompt Selection Frequency Regularization}, ensuring balanced utilization of prompts by discouraging overused ones and encouraging seldom-selected prompts, 
and (2) \textit{Prompt Knowledge Orthogonalization}, minimizing redundancy and fostering semantic diversity through reducing the cosine similarities between prompt pairs.
This mechanism can help the prompt pool capture a wide range of contextual cues, thereby improving the quality of the learned video representations.

To sum up, our main contributions are fourfold:
\begin{itemize}
    \item We propose a novel prompt learning framework, \textit{\textbf{EgoPrompt}}, tailored for egocentric action recognition. It explores the effect of semantic interplay in HOI understanding, addressing the unique challenges in egocentric video understanding.
    \item We introduce the \textit{\textbf{Unified Prompt Pool}} design, enabling cross-component interaction and capturing the HOI semantic information.
    \item We present a training objective, \textit{\textbf{Diverse Pool Criteria}}, to enhance prompt informativeness by encouraging balanced usage and enforcing semantic orthogonality.
    \item We conduct extensive experiments, including \textbf{within/cross-dataset and base-to-novel generalization}. Results consistently demonstrate the effectiveness of EgoPrompt in improving generalization capability.
\end{itemize}
    
\section{Related Work}
\label{sec:related work}

\subsection{Prompt Learning}
Prompt learning~\cite{doprompt,COOP,COCOOP,BiYao} has emerged as a lightweight and parameter-efficient alternative to traditional fine-tuning, enabling task-specific adaptations without modifying the entire backbone. 
Initially, prompt learning techniques were applied in the textual modality, where handcrafted prompts such as ``a photo of a [CLASS]'' were embedded in models like CLIP~\cite{CLIP} for zero-shot classification tasks. 
However, hand-crafted prompts often lack adaptability and explainability. 
To address these limitations, CoOp~\cite{COOP} introduced continuous prompts represented as trainable vectors appended to text tokens, while CoCoOp~\cite{COCOOP} further proposed image-conditioned prompts at the instance level to improve novel class generalization~\cite{chen2023category}.
To align the representations of dual-encoder architectures, PromptSRC~\cite{promptsrc} jointly tunes visual and textual encoders, employing regularization constraints between original and adapted representations. MaPLe~\cite{khattak2023maple} extends prompt learning to dual encoders by introducing deep prompts and a text-to-visual coupling module, effectively projecting textual knowledge into the visual encoder. Similarly, KgCoOp~\cite{kgcoop} and TCP~\cite{tcp} emphasize class-aware knowledge encoded in the text encoder to enhance generalization.
L2P~\cite{wang2022learning} models the task-specific knowledge with a prompt-pair form, providing a solution to catastrophic forgetting.
Due to its lightweight advantage, prompt learning has started to gain attention in EAR. For example, POV~\cite{xu2023pov} employs hand-object interaction (HOI) labels to guide optimization and uses exocentric datasets for auxiliary learning. However, the application of prompt learning to egocentric action recognition remains underexplored. 
In this work, we introduce prompt learning as a way to enhance the generalization capability of the egocentric model. By integrating techniques such as deep prompts and prompt pools, our proposed EgoPrompt framework enhances generalization capabilities while addressing the unique characteristics of egocentric video data.

\subsection{Egocentric Action Recognition}
The emergence of large-scale egocentric datasets such as Ego4D~\cite{ego4d}, EGTEA~\cite{egtea}, and Epic-Kitchens~\cite{epic-kitchen} has significantly advanced research in egocentric video understanding~\cite{pramanick2023egovlpv2egocentricvideolanguagepretraining,cheng2024egothinkevaluatingfirstpersonperspective}. These datasets provide diverse benchmarks for exploring egocentric perception, including action recognition~\cite{x-mic,AoP}, video summarization~\cite{egodistill,shiota2024egocentric,lin2022egocentricvideolanguagepretraining}, and object interaction understanding~\cite{JiRobobrain, robobrain2.0}.
To address the practical demands of egocentric action recognition, researchers have proposed various methods to enhance the efficiency and generalization of egocentric models. For instance, EgoDistill~\cite{egodistill} introduces IMU signals to reduce the computational cost of processing dense keyframes, while Ego-Only~\cite{ego-only} explores differences between egocentric and exocentric videos, employing a masked autoencoder~\cite{he2022masked} during pretraining to improve video understanding capabilities. Despite these advancements, most existing methods focus on intra-dataset evaluation, which limits their applicability to real-world scenarios where distribution shifts are prevalent.
X-MIC~\cite{x-mic} extends the generalization scenarios proposed in\cite{clipadapter, A5, vita-clip} for egocentric action recognition, including cross-dataset and base-to-novel generalization. Additionally, AoP~\cite{AoP} decouples egocentric action recognition into component-aware tasks, introducing an adapter-based approach that uses verb embeddings as prior knowledge to assist noun prediction under an open-vocabulary setting~\cite{chen2021eckpn}. While these approaches demonstrate the importance of component-aware modeling, they often neglect the constraints that nouns impose on verbs, which may hinder generalization performance.
Nonetheless, the potential of fully exploring component-based representation fusion remains untapped. Current methods fail to capture the semantic interplay between verbs and nouns, which is crucial for the goal-oriented nature (focusing on the HOI region) of egocentric actions. To address this challenge, we propose EgoPrompt, which incorporates a Unified Prompt Pool design to enable interaction between verb and noun component-specific representations.

\section{Methodology}
In this section, we introduce the technical details of our proposed \textbf{EgoPrompt}, a novel prompt learning approach specifically designed for the EAR task. 
Building on the foundations of the EAR
baseline (described in Sec.~\ref{preliminary}), EgoPrompt further facilitates the component knowledge
interaction (described in Sec.~\ref{EgoPrompt}), achieving a better understanding of egocentric semantic information.

\begin{figure*}[t]
  \centering
  \includegraphics[width=1\linewidth]{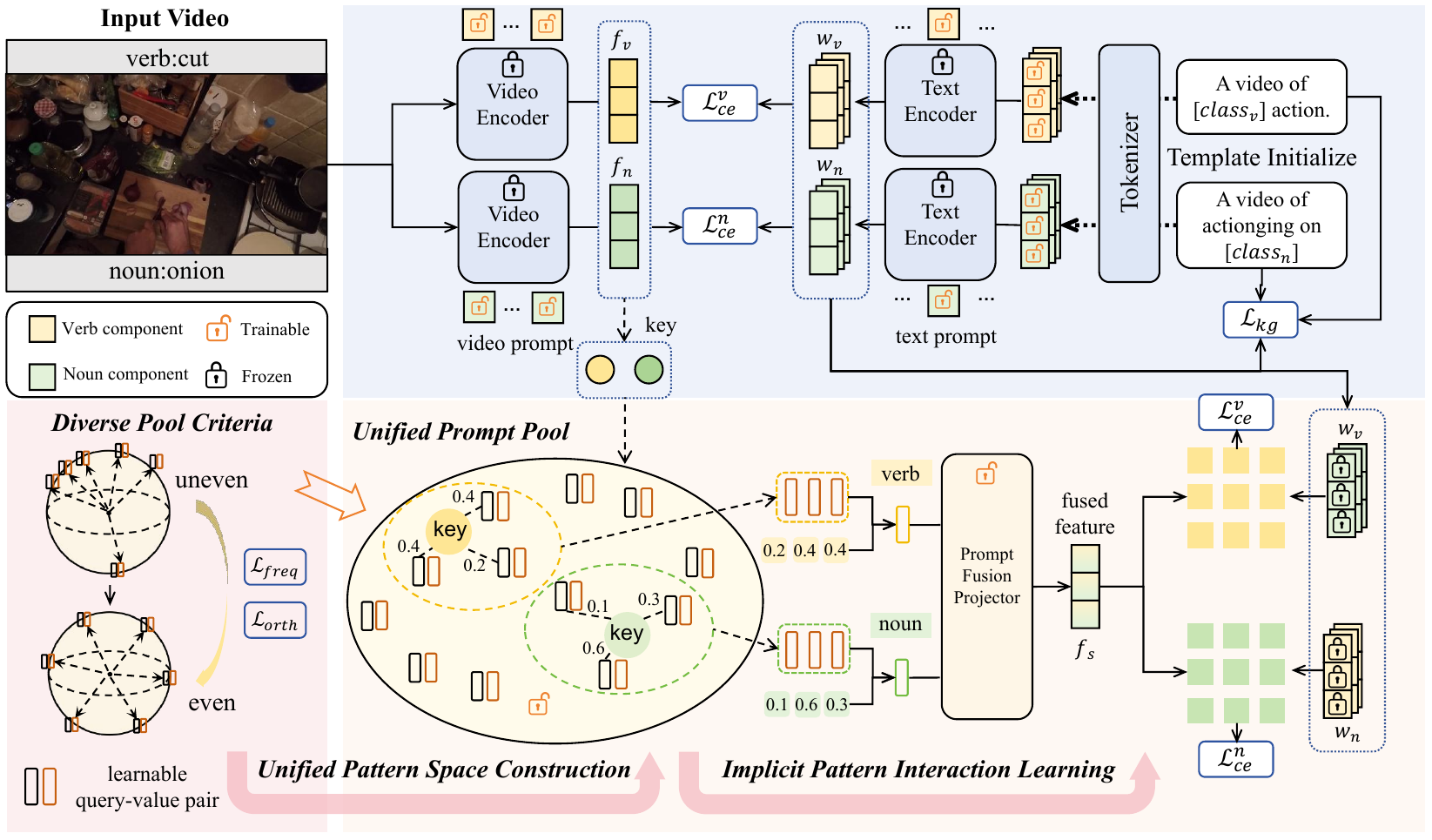}
  \caption{Overall framework of EgoPrompt. Building upon the Baseline work, EgoPrompt further establishes the semantic interaction between components. Specifically, under the guidance of the Diverse Pool Criteria, EgoPrompt constructs the Unified Pattern Space with the Unified Prompt Pool design. It decomposes the component-specific representation into fine-grained patterns and integrates the selected prompt pairs into a fused representation. }
  \label{framework}
\end{figure*}

\subsection{Baseline}
\label{preliminary}
In this work, we follow X-MIC~\cite{x-mic} to adopt LaVILA~\cite{lavila} as the backbone for egocentric action recognition.
LaVILA is a dual-encoder vision-language model that processes video and text inputs through separate transformer-based encoders.
To enhance cross-modal alignment, we incorporate the deep prompt learning strategy from MaPLe~\cite{khattak2023maple}. 
In this design, learnable textual prompts are inserted into each transformer layer of the text encoder. 
These prompts are then transformed into video-aware prompts via lightweight layer-wise projection modules and injected into the corresponding layers of the video encoder. 
For implementation details, please refer to MaPLe~\cite{khattak2023maple} and our supplementary material.

To adapt to the EAR task, we feed two sets of video prompts into the video encoders to extract component-specific representations $f_v$ and $f_n$ for verbs and nouns.
In the same way, the text encoder processes component-specific text prompts to generate the corresponding class embeddings $W^v, W^n$.
These text prompts are initialized using the tokenized results of our selected hand-crafted templates.
We empirically study the impact of template choice in the supplementary material and adopt the most effective ones, as illustrated in the Template Initialization stage of Fig.~\ref{framework}.
After this, both visual and textual prompts are jointly optimized using a component-specific classification loss: 
\begin{equation} 
\label{eq:ce}
\mathcal{L}_{ce}^{c} = \frac{\exp(\text{cos}(f_c, w_y^c)/\tau)}{\sum_{j=1}^{N^c} \exp(\text{cos}(f_c, w_j^c)/\tau)}, \quad c \in \{v, n\}, 
\end{equation} 
where $c$ denotes the component between verb and noun, and $w_y^c$ is the text embedding of the ground-truth class $y$ for component $c$.
$\text{cos}(\cdot)$ denotes cosine similarity, $\tau$ is a temperature factor, and $N^c$ is the number of classes for component $c$.

In addition, we apply a knowledge-guided loss~\cite{kgcoop} to preserve the semantic priors encoded in hand-crafted templates: 
\begin{equation} 
\label{eq:kg}
\mathcal{L}_{kg}^{c} = | W^c - \hat{W}^c |_2^2, \quad c \in \{v, n\}, 
\end{equation} 
where $\hat{W}^c$ denotes the class embedding obtained directly from the tokenized hand-crafted templates (without further training),
while $W^c$ is derived from text prompts that are initialized from these templates and then optimized during training.
This loss encourages the learned soft prompts to be semantically consistent with the original template.

\subsection{EgoPrompt}
\label{EgoPrompt}
As described above, the baseline lacks the ability to reason over the inherent correlations between actions and objects, which is essential for understanding egocentric HOI knowledge. 
Building upon the component-specific knowledge, we further develop \textbf{EgoPrompt} to facilitate the interaction within components. 
Specifically, EgoPrompt introduces two key modules: the \textit{Implicit Pattern Interaction Learning} and \textit{Unified Pattern Space Construction} strategy.

\subsubsection{Implicit Pattern Interaction Learning}
\label{implicit}

Although the component-specific prompting design effectively encodes component semantics, egocentric models still face challenges in irrelevant background noise and object clutter in egocentric scenes. 
Fortunately, the strong contextual association between verbs and nouns offers valuable cues for disambiguating such noisy scenes and identifying the core HOI knowledge. 
To fully exploit this semantic interplay, we introduce a \textit{\textbf{Unified Prompt Pool}}, which serves as a shared latent space for capturing implicit cross-component patterns. 
As illustrated in Fig.~\ref{framework}, this module enables dynamic interaction between the component-specific features by decomposing them into fine-grained patterns and re-integrating them into a unified representation.

The interaction process is achieved by utilizing the component-specific features $f_v$ (verb feature) and $f_n$ (noun feature) as keys to interact with query-value pairs in the Unified Prompt Pool. 
Specifically, the prompt pool contain $\bm{P}$ prompt pairs, where each pair consists of a query prompt $\bm{q}$ and a value prompt $\bm{v}$, denoted as: $\mathcal{P}=\{(\bm{q^1},\bm{v^1}),(\bm{q^2},\bm{v^2}),\cdots,(\bm{q^P},\bm{v^P})\}$.
The query prompt serves as the connection between the component-specific representation space and the unified prompt pool space, and the value prompt encodes the corresponding latent semantic pattern.
For each component (verb and noun), we first retrieve the Top-\(k\) most semantically similar query prompts based on the cosine similarity metric.
Then, we compute the attention weights $\alpha_i^c$ over the selected prompts using a temperature-scaled softmax:
\begin{equation}
\alpha_i^c = \frac{
\exp\left(\cos(f_c, \bm{q}_i)/\tau\right)
}{
\sum\limits_{\bm{q}_j \in \mathcal{Q}_c} \exp\left(\cos(f_c, \bm{q}_j)/\tau\right)
}, \quad \bm{q}_i \in \mathcal{Q}_c = \text{Top-}k(f_c, \mathcal{Q}),
\label{eq:topk_attention_simplified}
\end{equation}
where $\mathcal{Q}$ denotes the query prompt set.
Accordingly, we fetch the corresponding value prompts $\mathcal{V}_c=\{\bm{v_i|q_i}\in\mathcal{Q}_c\}$ and compute the pattern-composed feature for each component as:
\begin{equation} 
f_c' = \sum_{\bm{v}_i\in\mathcal{V}_c} \alpha_i^c \cdot \bm{v}_i, \quad c \in \{v,n\}.
\label{eq:attention_fusion} 
\end{equation}
After this, the final fusion representation is obtained by projecting the two pattern-based component representations:
\begin{equation}
f_{s} = \textbf{Proj}(f_v^{'},f_n^{'}),
\label{prompt fusion}
\end{equation}
where $\textbf{Proj}(\cdot)$ is a learnable projection layer.
This fusion representation $f_s$ encodes the implicit interaction between verbs and nouns, providing a noise-robust and component-shared embedding well aligned with egocentric HOI recognition.

\subsubsection{Unified Pattern Space Construction}
\label{unified}
To ensure that the prompt pool encodes a rich variety of informative patterns, we introduce the \textit{\textbf{Diverse Pool Criteria}}, a dedicated objective function designed to promote both effective prompt utilization and semantic diversity among prompt pairs.
This objective comprises two complementary regularization strategies:

1) \textit{Prompt Selection Frequency Regularization:}
To avoid prompt over-reliance and encourage balanced utilization across the pool, we reward the $k$ least-frequently selected prompts (encouraging more utilization) and penalize the $k$ most-frequently selected prompts (discouraging over-reliance). Let $c_p^c$ denote the selection count of prompt pair $p$ for component $c\in\{v,n\}$ during training. 
The regularization term is defined as:
\begin{equation}
\mathcal{L}_{\text{freq}} = - \sum_{c \in \{v,n\}}( \sum_{p \in \mathcal{S}_{\text{min}}} c_p^c - \sum_{p \in \mathcal{S}_{max}} c_p^c ),
\end{equation}
where $\mathcal{S}_{\text{min}}$ and $\mathcal{S}{_\text{max}}$ represent the sets of the $k$ least- and most-frequently selected prompts, respectively. 
$\lambda_{\text{freq}}$ is a weighting factor.
This regularization encourages underutilized prompts to participate more in learning and discourages prompt collapse into a narrow set of overused patterns.

2) \textit{Prompt Knowledge Orthogonalization:} To differentiate the knowledge encoded by prompt pairs, we apply an orthogonalization constraint over the query and value prompts in the pool. 
By minimizing the cosine similarity between prompt embeddings, we reduce redundancy and promote representation diversity.
The orthogonalization loss is defined as:
\begin{equation} 
\mathcal{L}_{\text{orth}} = \frac{1}{P(P - 1)}\sum_{i=1}^P \sum_{j=1, j \neq i}^P \left( \left| \frac{\bm{q}_i \cdot \bm{q}_j}{|\bm{q}_i||\bm{q}_j|} \right| + \left| \frac{\bm{v}_i \cdot \bm{v}_j}{|\bm{v}_i||\bm{v}_j|} \right| \right), 
\end{equation}
where $|\cdot|$ denotes the absolute value operator, $\bm{q}_i \cdot \bm{q}_j$ and $\bm{v}_i \cdot \bm{v}_j$ are the dot product between the $i$-th and $j$-th query and value prompts. 
This regularization $\mathcal{L}_{\text{orth}}$ encourages prompt pairs to occupy distinct subspaces, thereby enhancing the diversity of captured semantic patterns.
Overall, the final training objective for unified pattern space construction combines the cross-entropy losses from both verb and noun branches with the two regularization terms:
\begin{equation}
\mathcal{L}_{\text{uni}} = \mathcal{L}_{ce}^v + \mathcal{L}_{ce}^n + \lambda_{\text{freq}} \mathcal{L}_{\text{freq}} + \lambda_{\text{orth}} \mathcal{L}_{\text{orth}},
\label{overall loss}
\end{equation}
where $\lambda_{\text{freq}}$ and $\lambda_{\text{orth}}$ are balancing coefficients controlling the strength of the frequency and orthogonality regularizations, respectively.
We provide hyperparameter analysis in the supplementary materials.
By enforcing both usage balance and semantic orthogonality, the \textit{\textbf{Diverse Pool Criteria}} ensures that the unified prompt space remains rich, informative, and well-structured, supporting robust egocentric action recognition.
We summarize the complete training procedure in Algorithm~\ref{algorithm: EgoPrompt}.

\begin{algorithm}[tb]
\small
\caption{EgoPrompt Training Procedure}
\label{algorithm: EgoPrompt}
\begin{algorithmic}[1] 
\REQUIRE Component-specific prompts $\bm{p}_{v}, \bm{p}_{n}$; number of iterations $T_1, T_2$ for two training stages; Unified Prompt Pool $\mathcal{P}$ consisting of $P$ query-value prompt pairs $\{(\bm{q}^p, \bm{v}^p)\}_{p=1}^{P}$; Projector $\textbf{Proj}(\cdot)$.

\STATE \textbf{// Stage 1: Component-Specific Prompt Learning}
\FOR{$t = 1$ to $T_1$}
    \STATE Extract component-specific video features $f_v$, $f_n$.
    \STATE Obtain the classification loss $\mathcal{L}_{ce}$ and knowledge-guided $\mathcal{L}_{kg}$ loss for both components via Eq.~\ref{eq:ce} and Eq.~\ref{eq:kg}.
    \STATE Optimize $\bm{p}_v$ and $\bm{p}_n$.
\ENDFOR

\STATE \textbf{// Stage 2: Implicit Pattern Interaction Learning}
\FOR{$t = 1$ to $T_2$}
    \STATE Extract component-specific features $f_v$, $f_n$.
    \STATE Retrieve top-$k$ query prompts $\mathcal{Q}_v$, $\mathcal{Q}_n$ and their corresponding value prompts $\mathcal{V}_v$, $\mathcal{V}_n$ from pool $\mathcal{P}$ via Eq.~\ref{eq:topk_attention_simplified}.
    \STATE Compute fused representation $f_s$ via Eq.~\ref{eq:attention_fusion} and Eq.~\ref{prompt fusion}.
    \STATE Obtain the unified space construction objective $\mathcal{L}_{\text{uni}}$ via Eq.~\ref{overall loss}.
    \STATE Optimize the Unified Prompt Pool $\mathcal{P}$ and $\textbf{Proj}(\cdot)$.
\ENDFOR

\RETURN $\bm{p}_{v}, \bm{p}_{n}$, $\mathcal{P}$, and $\textbf{Proj}(\cdot)$.
\end{algorithmic}
\end{algorithm}

\section{Experiments}

\subsection{Datasets}
\textbf{Ego4D}~\cite{ego4d}. 
We follow X-MIC~\cite{x-mic} and use a subset of the Ego4D Forecasting Hands and Object (FHO) benchmark.
It is annotated with fine-grained noun and verb labels.
The training set consists of 64K video clips, with 521 noun classes and 117 verb classes.
The test set includes 33K clips with an average clip duration of 8 seconds, amounting to approximately 215 hours of video, excluding irrelevant background clips.

\noindent\textbf{Epic-Kitchens}~\cite{epic-kitchen} includes 67K video clips for training and 10K video clips for validation and testing. 
Each clip is 3.5 seconds on average, amounting to approximately 70 hours of video (irrelevant background clips are excluded). 
The dataset focuses on kitchen activities and is annotated for 300 noun classes and 97 verb classes.

\noindent\textbf{EGTEA}~\cite{egtea} is only used for testing generalization performance. 
Its training set includes 8,000 video clips with an average duration of 3.2 seconds, while the test split comprises 6,000 video clips. 
The dataset provides 20 verb classes and 54 noun classes.

\begin{table*}
\caption{\textbf{Comparison on the within- and cross-dataset generalization setting.} The superscript of $\oplus$ denotes the CoOp-based and $\odot$ denotes the egocentric-based algorithms., ``hm'' is short for harmonic average.} 
\label{CDG}
\centering
\scriptsize
\begin{tabular}{l|ccc|ccc|ccc|ccc}
\hline
 & \multicolumn{6}{c|}{Trained on Ego4D (E4D)} & \multicolumn{6}{c}{Trained on Epic-Kitchens (EK)} \\
 \cline{2-13}
           & \multicolumn{3}{c}{Nouns}                   & \multicolumn{3}{c|}{Verbs}                         &
             \multicolumn{3}{c}{Nouns}                   & \multicolumn{3}{c}{Verbs}                           \\
\hline
           & \multicolumn{1}{c}{E4D}                     & \multicolumn{1}{c}{EK}                              &
             \multicolumn{1}{c|}{hm}                      & \multicolumn{1}{c}{E4D}                            & 
             \multicolumn{1}{c}{EK}                       & \multicolumn{1}{c|}{hm}
           & \multicolumn{1}{c}{E4D}                     & \multicolumn{1}{c}{EK}                              &
             \multicolumn{1}{c|}{hm}                      & \multicolumn{1}{c}{E4D}                            & 
             \multicolumn{1}{c}{EK}                       & \multicolumn{1}{c}{hm}                             \\
\hline
CoOp{$^\oplus$}~\cite{COOP}        & 31.46 & 33.20 &  32.31                       & 22.50 & 23.08 & 22.79                               &
        14.80            &  34.70     &   20.75                              &  15.42     &  53.17     &               23.91                  \\
CoCoOp{$^\oplus$}~\cite{COCOOP}      & 35.71 & 33.06 &  34.33                       & 24.26 & 29.86 &26.77                                &
         12.76         &  36.93 &              18.97           & 14.28  & 55.20  &                         22.69        \\
CLIP-Adapter{$^\oplus$}~\cite{clipadapter}& 33.20 & 29.80 & 31.41                        & 23.68 & 31.92 & 27.19                           &
         11.40      &  37.85 &            17.52             & 16.13  & 56.77  &                              25.12   \\
A5{$^\oplus$}~\cite{A5}          & 34.58 & 31.42 & 32.92                        & 24.80 & 27.65 & 26.15                               &
        13.96    &  39.73 &        20.66                 &  16.79 & 56.05  &            25.84                     \\
Vita-CLIP{$^\oplus$}~\cite{vita-clip}   & 33.68 & 30.10 & 31.79                        & 23.46 & 28.90 & 25.90                               &
       16.47         &  40.08 &        23.35                 & 17.30  & 55.42  &                             26.37    \\
POV{$^\odot$}~\cite{xu2023pov}         & 37.60 & 32.86 &  35.07                       & 27.46 & 34.80 &  30.70                              & 13.75  & 37.70  &   20.15                      &  16.74 & 53.08  &                          25.45       \\
X-MIC{$^\odot$}~\cite{x-mic}       & 35.85 & 28.26 &  31.61                    & 28.27 &39.49  &   32.95           &
         11.45  & 44.07  &          18.17 &16.01            & 53.02  &       24.60                        \\
KgCoOp{$^\oplus$}~\cite{kgcoop}         & 36.71 & 34.18 &  35.40                       & 22.40 & 43.17 & 29.50                               &
           17.28       & 41.72  &          24.44               &  18.93 & 52.70  &                            27.85     \\
AoP{$^\odot$}~\cite{AoP}      & 34.60 & 34.26 & 34.43                        & 24.63 & 39.76 &  30.42                              &
          11.62          & 38.76  &           17.88              & 13.16  & 49.45  &                          20.79       \\
MaPLe{$^\oplus$}~\cite{khattak2023maple}       & 39.87 & 32.82 & 36.00                        & 25.53 & 44.24 &  32.38                              & 17.31   & 41.62  &            24.45             & 18.26  &  58.10 &              27.79                   \\ 
\hline
\rowcolor{blue!10}
EgoPrompt           & \textbf{42.93} & \textbf{35.75 }&\textbf{39.01}                         & \textbf{29.71} & \textbf{47.89} &\textbf{36.67}               &\textbf{19.45}    & \textbf{44.58}   & \textbf{27.08}          & \textbf{20.78}  &  \textbf{61.40} &             \textbf{31.05}              \\ 
\hline
\end{tabular}
\end{table*}

\begin{table*}
\caption{\textbf{Comparison on the base-to-novel class generalization setting.} The experiment results are pre-trained on Ego4D and evaluated on Epic-Kitchen. } 
\label{BN}
\centering
\scriptsize
\begin{tabular}{l|ccc|ccc|ccc|ccc}
\hline
 \multirow{3}{*}{\makecell{Pre-trained dataset: E4D\\Evaluation dataset: EK}}& \multicolumn{6}{c|}{Average Accuracy} & \multicolumn{6}{c}{Class Average Accuracy} \\
 \cline{2-13}
           & \multicolumn{3}{c}{Nouns}                   & \multicolumn{3}{c|}{Verbs}                         &
             \multicolumn{3}{c}{Nouns}                   & \multicolumn{3}{c}{Verbs}                           \\
\cline{2-13}
           & \multicolumn{1}{c}{base}                     & \multicolumn{1}{c}{novel}     & \multicolumn{1}{c|}{hm}                                       & \multicolumn{1}{c}{base}                            & 
             \multicolumn{1}{c}{novel}  & \multicolumn{1}{c|}{hm}                       
           & \multicolumn{1}{c}{base}                     & \multicolumn{1}{c}{novel}   & \multicolumn{1}{c|}{hm}                                                 & \multicolumn{1}{c}{base}                            & 
             \multicolumn{1}{c}{novel}       & \multicolumn{1}{c}{hm}                                             \\
\hline
CoOp{$^\oplus$}       & 34.16 & 23.18    &   27.62                 & 24.27 & 2.04     &     3.76                   &
                20.05    & 5.18    &   8.23                 & 10.07 & 1.40    &   2.46                           \\
CoCoOp{$^\oplus$}     & 35.17 & 23.07     &   27.86                & 24.63 & 1.52    &     2.86                    &
                20.98    & 5.71  & 8.98 &        11.63                &   1.71 & 2.98                            \\
CLIP-Adapter{$^\oplus$}  &33.28  & 21.72       &   26.29            & 30.74 & 2.71           &    4.98              &
                19.76    & 4.98  & 7.96 &     11.92                &  2.04 &3.48                          \\
A5{$^\oplus$}          & 34.10 & 22.90          &   27.40          & 27.60 & 2.80            &    5.08
                & 20.73 & 3.48   &5.96&       10.37        &   1.82 &3.10                           \\
Vita-CLIP{$^\oplus$}  &  32.18  & 22.67 &          26.60       &      29.14 & 2.63           &    4.82               &
                  19.64  & 3.70  &6.23&        11.31            & 1.62&2.83                           \\
POV{$^\odot$}        &   34.95 & 22.17                        & 27.13 &   33.85              &    2.09
                    & 3.94  &   20.85   & 4.07     &6.81              &   11.68   &     1.79 &3.10                      \\
X-MIC{$^\odot$}       & 34.32  & 23.00               & 27.54 &    30.17                          &  2.42
                    & 4.48  &   20.46  &5.02       & 8.06              &     10.80&   1.93  &3.27                           \\
AoP{$^\odot$}        &36.33 & 24.01                        &  28.91&    41.52                     &  2.13   &   4.05
                   &  21.61  &  6.80    & 10.34    &12.30 & 2.60   &4.29                                            \\
MaPLe{$^\oplus$}      &35.70 & 24.13        &      28.80      & 46.09 & 2.68            &       5.07           &
                    21.13 & 6.32  &9.73           &          12.76& 2.15&3.68                                \\ 
                    \hline
                   \rowcolor{blue!10}
EgoPrompt           & \textbf{36.91}  & \textbf{24.34}         &    \textbf{29.34}     & \textbf{50.70} & \textbf{3.07}            &     \textbf{5.79}           &
                    \textbf{21.83}  & \textbf{7.14}     & \textbf{10.76}     & \textbf{13.77} & \textbf{3.21}  &  \textbf{5.21}                    \\ 
\hline
\end{tabular}
\vspace{-1.2em}
\end{table*}

\subsection{Expermiental Setup}

\noindent\textbf{Training Details.} Our implementation is adapted from the public code of the X-MIC~\cite{x-mic}. 
To make a fair comparison, we adopt dual-encoder architecture based on the LAVILA~\cite{lavila} with \textbf{Timesformer-Large} as our default backbone for all algorithms. 
The prompt’s length is set to 4 for both video, textual modalities, and prompt pool construction. 
For initialization, we employ hand-crafted templates in the form of ``a video of a [CLASS] action'' for verb classes and ``a video of actioning on [CLASS]'' for noun classes. 
The Adamw~\cite{AdamW} optimizer is applied for optimization with the learning rate of 1e-4 for formal training and 2e-5 for warmup, and the batch size in Ego4D and Epic-Kitchen is set for 64 and 32, and training epochs is 5, including 3 warm-up epochs. 
All experiments are conducted on 4$\times$A800 GPUs with a single training run lasting 8 hours.

\noindent\textbf{Baselines.} To introduce the prompt-learning technique to egocentric action recognition, we re-implement several recent CoOp-based works, \emph{e.g.}, MaPLe~\cite{khattak2023maple} and KgCoOp~\cite{kgcoop}.
In addition, we introduce some state-of-the-art egocentric action recognition methods, X-MIC~\cite{x-mic}, AoP~\cite{AoP}, and POV~\cite{xu2023pov}. 
To make a comprehensive comparison, we also include results from conventional supervised fine-tuning baselines, as illustrated in Fig.~\ref{figure:DG} and Fig.~\ref{figure:adapt}.  
All these methods are evaluated under consistent protocols on the following proposed benchmarks.

\noindent\textbf{Benchmarks.}
We evaluate EgoPrompt under two standard egocentric action recognition benchmarks:
\textit{1) Within- and Cross-dataset generalization:} 
In this setting, models are trained on a single dataset and evaluated on both the training domain (within-dataset) and an unseen testing domain (cross-dataset).
This benchmark assesses the model’s ability to not only fit the source domain but also to generalize to new environments.
Accordingly, we report performance on both domains to reflect the model's adaptability to distribution shifts.
\textit{2) Base-to-novel class generalization:} Here, datasets are divided into base and novel subsets based on whether the data label is shared across training and testing datasets or appears exclusively in the testing dataset. 
The model is trained solely on the base class subset of the training dataset 
and evaluated on both the base and novel classes within the testing dataset. 
Performance on base classes indicates the model’s ability to transfer category-level knowledge across domains, while performance on novel classes reflects its zero-shot recognition capability.

\noindent\textbf{Evaluation Metrics.} 
To assess generalization fairly, we adopt two complementary metrics:
\textit{Average Accuracy}: Measures the overall accuracy across all test samples, capturing instance-level correctness.
\textit{Class Average Accuracy}: Computes the mean accuracy across classes, mitigating bias from class imbalance and emphasizing performance consistency on both frequent and rare classes.
Together, these metrics provide a holistic view of the model’s generalization behavior from both sample-level and class-level perspectives.

\subsection{Within- and Cross-Dataset Generalization}
\noindent\textbf{Prompt Learning \textit{vs.} Fine-tuning \textit{vs.} Zero-shot.}
As shown in Fig.~\ref{figure:DG}, prompt-based approaches (e.g., X-MIC and EgoPrompt) significantly outperform both zero-shot CLIP and fully fine-tuned (SFT) baselines in cross-dataset generalization. 
It is because zero-shot models suffer from poor alignment to downstream distributions, and SFT tends to overfit to the source domain, while prompt learning can help achieve a better balance between adaptability and generalization.
Notably, EgoPrompt achieves the highest performance across both datasets and categories, especially in verb classification. For example, on Epic-Kitchens, EgoPrompt improves verb accuracy from 5.3\% (zero-shot) and 16.8\% (SFT) to 47.9\%, clearly demonstrating the effectiveness of structured prompt learning in capturing transferable action semantics.
Compared to X-MIC, which also introduces learnable modules, EgoPrompt further enhances generalization by modeling cross-component interactions and latent semantics, leading to consistent gains on both nouns and verbs.

\noindent\textbf{Ego4D \& EpicKitchen.} 
We present a detailed comparison of generalization performance in Table~\ref{CDG}, where models are trained on the source dataset and tested on both the source (i.e., within-dataset) and target datasets (i.e., cross-dataset) for noun and verb classification.
(1) Within-dataset performance: EgoPrompt achieves the highest within-dataset accuracy on E4D (42.93\% for nouns and 29.71\% for verbs) and EK (44.58\% for nouns and 61.40\% for verbs). These results exceed the prior best-performing method MaPLe by +3.06\% and +4.18\% in E4D and +2.96\% and +3.30\% in EK, showing that our method not only generalizes well but also retains strong fitting ability on the within dataset.
(2) Cross-dataset performance: EgoPrompt also outperforms all baselines on the unseen test set, reaching 35.75\% noun and 47.89\% verb in EK testing and 19.45\% noun and 20.78\% verb in E4D testing. These consistent improvements are also reflected in the hm, with EgoPrompt outperforming MaPLe by +3.65\% nouns and +4.29\% verbs in EK testing and +2.14\% nouns and 2.52\% verbs in E4D testing.
As a result, the experimental results demonstrate the effectiveness of EgoPrompt in both within- and cross-domain generalization gains.

\noindent\textbf{Ego4D \& EGTEA.} 
To further evaluate the cross-dataset generalization capability of EgoPrompt, we report its performance on the EGTEA dataset in Table~\ref{DG}, where models are trained on Ego4D and tested directly on EGTEA without any adaptation. EgoPrompt achieves the best performance across all metrics, reaching 32.8\% on noun classification and 40.3\% on verb classification, surpassing all baselines by a clear margin.
Compared to MaPLe, which previously achieved 30.7\% (nouns) and 36.2\% (verbs), EgoPrompt improves performance by +2.1\% and +4.1\%, respectively. These improvements are further reflected in the hm, where EgoPrompt achieves 36.2\%, outperforming MaPLe (33.2\%) and other methods, including X-MIC (30.3\%) and KgCoOp (32.0\%).
Together with the results on Epic-Kitchens, these findings demonstrate that EgoPrompt not only retains strong within-domain and base class performance but also generalizes robustly to unseen domains and novel classes, effectively bridging the gap across egocentric video datasets.

\begin{figure}[t]
    \centering
    \includegraphics[width=1.0\columnwidth]{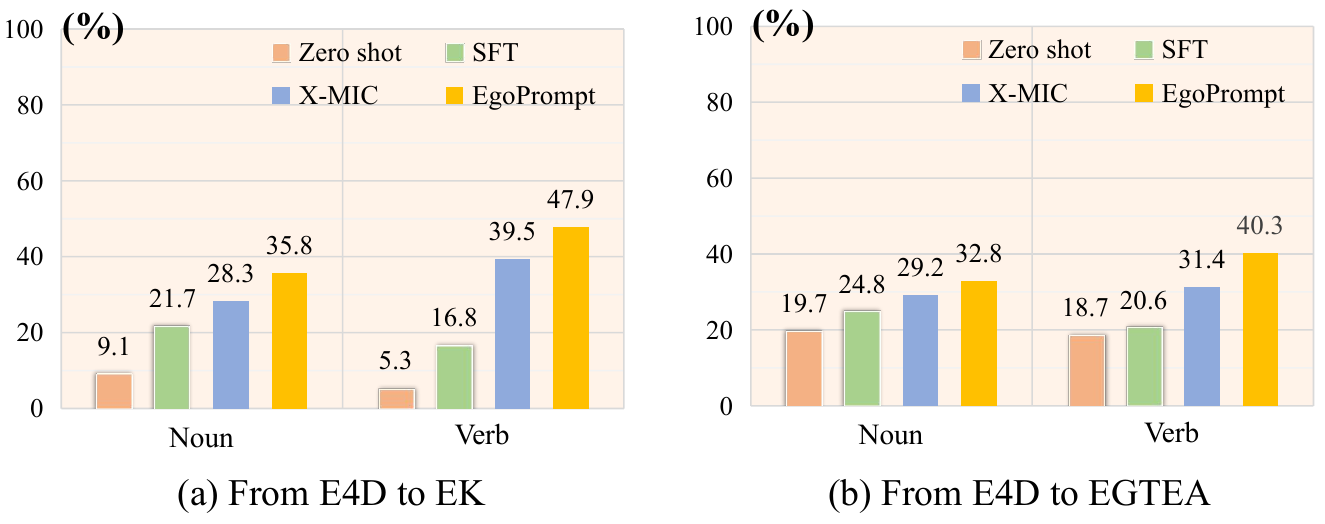}
    \caption{\textbf{Comparison of the generalization performance.} The sub-caption denotes the training and testing dataset in this cross-dataset generalization setting}
    \label{figure:DG}
\end{figure}

\subsection{Base-to-Novel Class Generalization}
In Table~\ref{BN}, we evaluate the generalization performance of various methods under the base-to-novel setting, where models are trained on base classes in Ego4D and tested on both base and novel classes in Epic-Kitchens.
EgoPrompt demonstrates the strongest performance across all metrics. It achieves substantial gains on base class recognition, particularly for verbs, reaching 50.70\% in average accuracy—surpassing CoOp (24.27\%) and MaPLe (46.09\%) by a large margin. This highlights EgoPrompt’s ability to extract more transferable action semantics in commonly occurring categories.
In novel classes, all methods exhibit limited performance due to the inherent challenge of few-shot generalization. EgoPrompt slightly improves novel verb accuracy to 3.07\%, compared to 2.04\% of CoOp and 2.68\% of MaPLe. Although this marks the highest result among all baselines, the absolute value remains low, indicating that the egocentric model still struggles in long-tail scenarios with scarce training data.
From a class-level perspective (Class Average Accuracy), EgoPrompt again leads on both nouns and verbs across base and novel splits. Notably, the hm for verbs improves to 5.21\%, which is +1.53\% higher than MaPLe, confirming the consistent gains of our method.
In summary, EgoPrompt excels in modeling base class semantics and offers modest gains on novel classes.

\begin{table}
\caption{\textbf{Cross-dataset generalization on EGTEA.}} 
\vspace{-0.8em}
\label{DG}
\centering
\small
\begin{tabular}{l|ccc|ccc}
\hline
 & \multicolumn{3}{c|}{E4D (within)} & \multicolumn{3}{c}{EGTEA (cross)}  \\
\cline{2-7}
 & \multicolumn{1}{c}{nouns}                              & \multicolumn{1}{c}{verbs}                                                 & \multicolumn{1}{c|}{hm}                               & \multicolumn{1}{c}{nouns}                                                 & \multicolumn{1}{c}{verbs}                              & \multicolumn{1}{c}{hm}                                \\
\hline
X-MIC      &35.9&29.0 & 32.5 & 29.2& 31.4&30.3 \\
KgCoOp     &36.7&22.4 & 27.8 & 29.8& 34.6&32.0 \\
AoP        &34.6&24.6 & 28.8 & 30.1& 32.6&31.3 \\
MaPLe      &39.9&25.5 & 31.1 & 30.7& 36.2&33.2 \\
\hline
\rowcolor{blue!10}
EgoPrompt  &\textbf{42.9}&\textbf{29.7} & \textbf{35.1} &\textbf{32.8}&\textbf{40.3}&\textbf{36.2}\\
\hline
\end{tabular}
\vspace{-1.2em}
\end{table}

\subsection{Ablation Study}
EgoPrompt combines prompt learning with the characteristics of egocentric action recognition tasks. 
Therefore, we conducted a series of ablation studies to validate the effectiveness of the EgoPrompt.
All of our ablation experiments are pre-trained on Ego4D.

\noindent\textbf{Effect of EgoPrompt training stage design.} 
EgoPrompt adopts a two-stage training strategy to balance component-specific representation learning and cross-component interaction. Specifically, Stage 1 focuses on Dual-Branch Prompt Learning, while Stage 2 captures implicit pattern interactions through the Unified Prompt Pool.
To evaluate the contribution of each stage and the training strategy, we conduct an ablation study as shown in Table~\ref{module}. We consider four variants: training with only Stage 1, only Stage 2, joint training of Stage 1 and 2 from scratch (denoted as ``Stage 1+2''), and the full two-stage training used in EgoPrompt.
Results show that removing either stage leads to notable performance degradation. Training with only Stage 1 yields 31.28\% (nouns) and 41.92\% (verbs) on Epic-Kitchens, while only Stage 2 gives slightly higher scores at 33.15\% and 46.30\%, indicating that interaction modeling contributes more to cross-domain generalization.
Interestingly, jointly training both stages from scratch (``Stage 1+2'') fails to reach the same performance as our two-stage strategy, dropping to 32.18\% (nouns) and 44.20\% (verbs). This confirms that sequential optimization—first learning disentangled component-specific prompts and then modeling their interaction—is more effective for robust and transferable egocentric representations.
These results validate the necessity of the proposed two-stage paradigm in EgoPrompt for achieving optimal performance.

\begin{figure}[t]
    \centering
    \includegraphics[width=1.0\columnwidth]{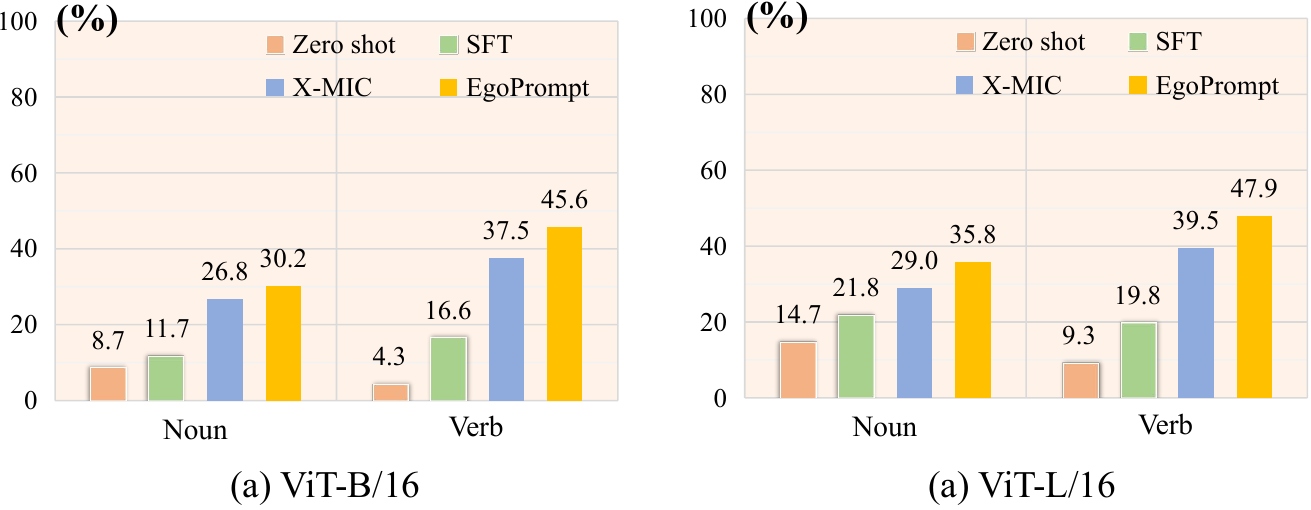}
    \vspace{-1.6em}
    \caption{\textbf{Adaptability of EgoPrompt on different backbones.} The above results are collected from cross-dataset generalization, ``From E4D to EK'' setting.}
    \label{figure:adapt}
   \vspace{-0.8em}
\end{figure}

\begin{table}[t]
\caption{\textbf{Effect of modules in EgoPrompt.}} 
\vspace{-0.8em}
\label{module}
\centering
\small
\begin{tabular}{ccc|cc|cc}
\hline
\multirow{2}{*}{Stage 1} & \multirow{2}{*}{Stage 2} & \multirow{2}{*}{Stage 1+2} & \multicolumn{2}{c|}{E4D (within)} & \multicolumn{2}{c}{EK (cross)} \\ 
\cline{4-7}
                     &                      &                      & nouns & verbs & nouns & verbs\\ \hline

\checkmark         & \mbox{-} & \mbox{-} &42.17 & \textbf{29.82} & 31.28 & 41.92\\ 
    \mbox{-}     & \checkmark &  \mbox{-} &40.18 & 28.31 & 33.15 & 46.30 \\ 
    \mbox{-}     & \mbox{-} &  \checkmark & 40.65 &27.90 & 32.18 & 44.20   \\ \hline
\rowcolor{blue!10}
\checkmark         & \checkmark & \mbox{-} &\textbf{42.93}&29.71 &\textbf{35.75}& \textbf{47.89} \\
\hline
\end{tabular}
\vspace{-1.2em}
\end{table}

\begin{figure*}[t]
    \centering
    \includegraphics[width=2.0\columnwidth]{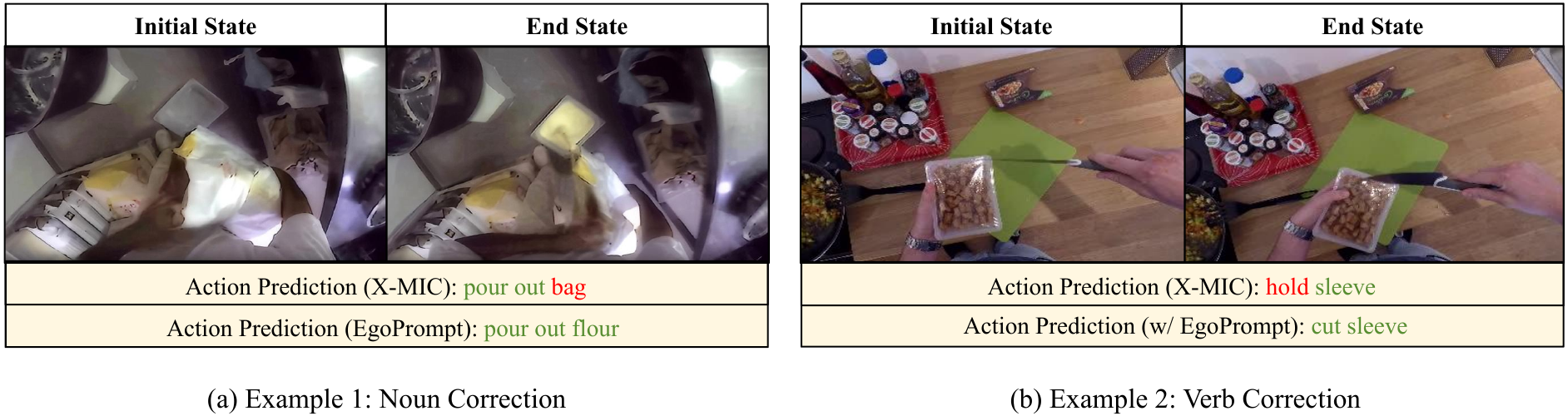}
    \vspace{-0.8em}
    \caption{\textbf{Qualitative examples of EgoPrompt’s improvements.} 
    (a) Noun correction: The baseline (X-MIC) incorrectly grounds the object to ``bag,'' while EgoPrompt leverages verb-centric semantics (e.g., deformable nature of ``pour out'') to infer the correct noun ``flour.'' 
    (b) Verb correction: The baseline misinterprets the action as ``hold sleeve.'' EgoPrompt captures the state change of the noun (i.e., sleeve being cut), which contradicts the static nature of ``hold'' and corrects the verb to ``cut.'' 
    Green and red highlights indicate correct and incorrect predictions, respectively.
    }
    \label{example1}
   \vspace{-0.8em}
\end{figure*}

\noindent\textbf{Adaptability across backbone scales.} 
To assess the scalability and adaptability of EgoPrompt under varying model capacities, we evaluate its performance on two video backbones: ViT-B/16 and ViT-L/16. The results are illustrated in Fig.~\ref{figure:adapt}.
Across both backbones, EgoPrompt consistently outperforms zero-shot CLIP, supervised fine-tuning (SFT), and the adapter-based method X-MIC for both noun and verb recognition. 
On ViT-B/16, EgoPrompt achieves 30.2\% (nouns) and 45.6\% (verbs), exceeding X-MIC by +3.4\% and +8.1\%, respectively. When scaled up to ViT-L/16, the performance further improves to 35.8\% (nouns) and 47.9\% (verbs), showcasing enhanced generalization with increased backbone capacity.
These results demonstrate that the proposed EgoPrompt is not only effective but also robust and adaptable across different model scales.

\noindent\textbf{Effect of Constraints in Diverse Pool Criteria.}
Table~\ref{module} presents an ablation study on the two regularization terms of the Diverse Pool Criteria: Prompt Selection Frequency Regularization ($\mathcal{L}_{freq}$) and Prompt Orthogonalization ($\mathcal{L}_{orth}$). Two terms are designed to encourage a more balanced and diverse utilization of prompt pairs in the Unified Prompt Pool, respectively.
Without either constraint, the model exhibits suboptimal generalization, particularly on cross-dataset (EK) evaluation, achieving only 34.46\% in hm. Adding $\mathcal{L}_{freq}$ alone leads to a clear improvement (hm +3.53\% on EK), indicating that preventing over-reliance on a subset of prompt pairs helps improve verb classification in particular.
Similarly, incorporating $\mathcal{L}_{orth}$ alone also boosts performance (hm +5.43\% on EK), especially on noun recognition, by encouraging prompt diversity and reducing redundancy within the pool. When both regularizations are applied jointly, the model achieves the highest performance across all metrics, with 35.12\% hm on E4D and 40.94\% on EK. 
These results confirm the complementary benefits of frequency balancing and orthogonal diversity, which together enhance the robustness and generalizability of the learned prompt space.

\begin{table}[t]
\caption{\textbf{Effect of Components in Diverse Pool Criteria.}} 
\vspace{-0.8em}
\label{module}
\centering
\small
\begin{tabular}{cc|ccc|ccc}
\hline
\multirow{2}{*}{$\mathcal{L}_{freq}$} & \multirow{2}{*}{$\mathcal{L}_{orth}$} & \multicolumn{3}{c|}{E4D (within)} & \multicolumn{3}{c}{EK (cross)} \\ 
\cline{3-8}
                     &                                        & nouns & verbs & hm & nouns & verbs & hm\\ \hline
    \mbox{-}     &\mbox{-}  & 38.31 & 24.71 &30.04 & 30.17 & 40.16 &34.46 \\ 
\checkmark& \mbox{-}   & 41.82& 27.64 &33.28 & 32.10 & 46.52 &37.99 \\ 
   \mbox{-}      &    \checkmark & 40.96 & 28.54 &33.64  & 34.60 & 47.08 &39.89 \\ \hline
\rowcolor{blue!10}
\checkmark   & \checkmark  &\textbf{42.93}&29.71 &35.12&\textbf{35.75}& \textbf{47.89} & 40.94\\
\hline
\end{tabular}
\end{table}


\noindent\textbf{Effect of Unified Prompt Pool size.}
Table~\ref{templates} investigates the impact of the prompt pool size ($\bm{P}$) on model performance. A smaller pool size (e.g., $\bm{P}=4$) provides limited pattern diversity, resulting in lower accuracy, particularly for nouns (hm = 33.88\%). As the pool size increases, performance improves across both within- and cross-dataset settings. The best results are obtained when $\bm{P}=16$, yielding the highest harmonic mean for both noun (35.12\%) and verb (40.94\%) classification.
Interestingly, further enlarging the pool to $\bm{P}=32$ leads to a slight performance drop, suggesting that overly redundant prompts may dilute the selection quality and weaken discriminative learning. These findings indicate that a moderate prompt pool size strikes a favorable trade-off between pattern diversity and prompt effectiveness. In particular, setting $\bm{P}=16$ enables the model to capture rich yet manageable semantic patterns for more robust cross-domain generalization. More Results can be found in the supplementary materials.

\subsection{Qualitative Analysis}
We present two representative examples in Figure~\ref{example1} to highlight how EgoPrompt enhances the model’s ability to reason over separate verb-noun semantics in egocentric videos.
In \textbf{Example~1} (Fig.~\ref{example1} (a)), the baseline model mispredicts the action as ``pour out bag,'' mistakenly linking the acted object to the ``bag.'' 
EgoPrompt captures the verb-centric semantic attribute of ``pour out,'' which typically involves a deformable or flowable object, successfully excluding the ``bag'' as a candidate.
Based on this, EgoPrompt corrects the prediction to the right answer.
In \textbf{Example~2} (Fig.~\ref{example1} (b)), the baseline incorrectly labels the action as ``hold sleeve,'' overlooking the dynamic interaction occurring in the scene.
In contrast, EgoPrompt leverages the object state change of the ``sleeve'' (being cut off), which is encoded in the noun representation.
The cues contained in noun representation conflict with the semantics of ``hold,'' which implies a stable and static object state.
By utilizing this inconsistency, EgoPrompt effectively rules out ``hold'' as a plausible verb and instead predicts ``cut sleeve,'' which better aligns with the observed scene dynamics.

\begin{table}
\caption{\textbf{Effect of Prompt Pool Size.}} 
\vspace{-0.8em}
\label{templates}
\centering
\small
\begin{tabular}{@{}c|ccc|ccc@{}}
\hline
\multirow{2}{*}{Pool Size $\bm{P}$}           & \multicolumn{3}{c|}{Nouns}                  & \multicolumn{3}{c}{Verbs}                 \\
\cline{2-7}
           & \multicolumn{1}{c}{E4D}                     & \multicolumn{1}{c}{EK}                        &
             \multicolumn{1}{c|}{hm}                      & \multicolumn{1}{c}{E4D}                         & 
             \multicolumn{1}{c}{EK}                       & \multicolumn{1}{c}{hm}                         \\
\hline
    4          & 41.60 & 28.58& 33.88                      & \textbf{36.14} & 46.50 & 40.67                          \\
   8  & 42.32  & \textbf{30.00}  & 35.11                        & 34.98 & 47.20 & 40.18              \\ \rowcolor{blue!10}
 16 & \textbf{42.93}&\textbf{29.71}& \textbf{35.12}&\textbf{35.75}& \textbf{47.89} &\textbf{40.94}                       \\
 32 & 42.70 & 29.11 &34.62      & 33.26 & 45.84 &38.55  \\
\hline
\end{tabular}
\end{table}

\section{Conclusion}
In this paper, we propose EgoPrompt, a novel prompt learning framework tailored for egocentric action recognition. 
EgoPrompt explicitly models the component interaction through a Unified Prompt Pool, enabling the construction of a more semantically aligned and context-aware fused representation.
To further enhance the effectiveness and diversity of learned prompt pairs, we introduce the Diverse Pool Criteria, which encourages balanced usage and orthogonal semantics across prompts. 
Extensive experiments demonstrate the superior performance of EgoPrompt. 

\noindent\textbf{Discussion.} While EgoPrompt improves generalization and interpretability in egocentric recognition, challenges such as the long-tail distribution remain. 
Future research could explore incorporating temporal attention or dynamic prompt adaptation strategies, extending the temporal HOI understanding capability of EgoPrompt.

\section{Acknowledgment}
This work was supported by the National Natural Science Foundation of China under Grants 62036012, U23A20387, 62322212, in part by the Pengcheng Laboratory Research Project under Grant PCL2023A08, and also in part by the Postdoctoral Fellowship Program of CPSF under Grant Number GZC20251036.



\bibliographystyle{ACM-Reference-Format}
\bibliography{main}
\end{document}